# About Updating.

**Philippe Smets**
IRIDIA, Université Libre de Bruxelles
50 av. F. Roosevelt, CP194/6. B-1050 Brussels, Belgium.

**Abstract:**
Survey of several forms of updating, with a practical illustrative example.

We study several updating (conditioning) schemes that emerge naturally from a common scenario to provide some insights into their meaning. Updating is a subtle operation and there is no single method, no single 'good' rule. The choice of the appropriate rule must always be given due consideration. Planchet (1989) presents a mathematical survey of many rules. We focus on the practical meaning of these rules. After summarizing the several rules for conditioning, we present an illustrative example in which the various forms of conditioning can be explained.

## 1. CONDITIONING RULES FOR BELIEF FUNCTIONS.

Let $\Omega$ be a finite set with elements $\omega_1, \omega_2, ...\omega_n$. The $\omega_i$ are mutually exclusive possible answers to a given question. Let bel: $2^\Omega \rightarrow [0,1]$ be a belief function over $\Omega$, with m its corresponding basic belief mass assignment. For $A \subseteq \Omega$, bel(A) quantifies our degree of belief that the true answer to the question is in A. If the $\omega_i$ are an exhaustive list of possible answers, the closed-world assumption prevails (as no solution exists outside $\Omega$), otherwise the open-world assumption prevails (Smets 1988)

Suppose a new piece of evidence that says that the answer is not in $\overline{A}$, where $\overline{A}$ is the set of $\omega_i$ that are elements of $\Omega$, but not of A. To say that the answer is not in $\overline{A}$ does not mean that the answer is in A. Under closed-world assumption both statements are equivalent. Under open-world assumption, they are not. Here, the answer might be none of the elements of $\Omega$. When I say "the answer is not in $\overline{A}$", I only eliminate the elements of $\overline{A}$ as possible answer. When I say "the answer is in A", not only I eliminate the elements of $\overline{A}$, but I also claim that the answer is an element of A - a much stronger claim. This distinction is at the origin of the difference between the open-world and the closed-world assumptions.

The updating of bel induced by the information that the answer is not in $\overline{A}$ can be performed variously. It will depend on the interpretation of the problem that bel is supposed to model.

### Conditioning C.1. The Unnormalized Dempster's Rule of Conditioning.

For all $X \subseteq \Omega$, the basic belief masses m(X) given to X is transferred to $A \cap X$. The basic belief masses assignment $m_A$, the belief function $bel_A$ and the plausibility function $pl_A$ obtained after conditioning on A are:

$$m_A(B) = \sum_{X \subseteq \overline{A}} m(B \cup X) \qquad \text{for } B \subseteq A$$

$$m_A(B) = 0 \qquad \text{for } B \not\subseteq A$$

$$bel_A(B) = bel(B \cup \overline{A}) - bel(\overline{A}) \quad \text{for } B \subseteq \Omega$$

$$pl_A(B) = pl(B \cap A) \qquad \text{for } B \subseteq \Omega$$

This solution is always applicable, even when pl(A) = 0. Notice that $m_A(\emptyset)$ might be non null with this solution. In fact, one has: $m_A(\emptyset) = m(\emptyset) + bel(\overline{A})$.

This rule is the one described in the transferable belief model under open world assumption (Smets 1988).

### Conditioning C.2. The Normalized Dempster's Rule of Conditioning.

Masses are transferred as in conditioning C.1, but the result is then proportionally normalized to cope with the masses that would be transferred to the empty set. This avoids ending with a positive mass on $\emptyset$, and guarantees that bel($\Omega$) = 1. After conditioning, one gets:

$$m_A(B) = c \sum_{X \subseteq \overline{A}} m(B \cup X) \qquad \text{for } B \subseteq A, B \neq \emptyset$$

$$m_A(B) = 0 \qquad \text{for } B \not\subseteq A$$

and for $B \subseteq \Omega$
$$bel_A(B) = c (bel(B \cup \overline{A}) - bel(\overline{A}))$$
$$pl_A(B) = c \, pl(B \cap A)$$
where $c^{-1} = bel(\Omega) - bel(\overline{A}) = pl(A)$.

This solution applies only when pl(A)>0. No solution is provided when pl(A)=0.

This rule is the one described in the transferable belief model under closed world assumption. It is the one initially proposed by Shafer (1976a).

**Conditioning C.2'. Bayesian solution.**

The classical solution with probability function is,

$$P_A(B) = \frac{P(B \cap A)}{P(A)} \quad \text{for } B \subseteq \Omega.$$

This solution applies only if P(A)>0. If P(A)=0, no solution is provided.

This is a particular case of conditioning C.2. It is obtained when the belief function bel is a probability function.

**Conditioning C.3. Yager-Kohlas's Solution.**

The basic belief masses are transferred as in C.1, but the masses that could be transferred to the empty set are reallocated to A, so:

$$m_A(B) = \sum_{X \subseteq \overline{A}} m(B \cup X) \quad \forall B \subseteq A, B \neq A, B \neq \emptyset$$

$$m_A(A) = \sum_{X \subseteq \overline{A}} m(A \cup X) + bel(\overline{A})$$

$$m_A(B) = 0 \quad \forall B \not\subseteq A$$

$$bel_A(B) = bel(B \cup \overline{A}) - bel(\overline{A}) \quad \forall B \subseteq A, B \neq A$$

$$bel_A(A) = bel(\Omega)$$

$$pl_A(B) = pl(B \cap A) + bel(\overline{A}) \quad \forall B \subseteq A$$

This conditioning might seem artificial, but it will be shown that it can be observed sometimes. It was proposed in Yager (1985) and Kohlas (1989). It applies normally only to normalized belief functions (i.e., those with m(∅)=0 or equivalently with bel(Ω)=1)

**Conditioning C.4. Geometric Rule of Conditioning.**

One way to condition belief functions consists in deciding that all basic belief masses not given to subsets of A are nullified, and those given to subsets of A are proportionally normalized so their sum remains one. Then $m_A$, $bel_A$ and $pl_A$ become:

$$m_A(B) = \frac{m(B)}{bel(A)} \quad \text{if } B \subseteq A$$
$$= 0 \quad \text{otherwise}$$

$$bel_A(B) = \frac{bel(B \cap A)}{bel(A)} \quad \forall B \subseteq \Omega$$

$$pl_A(B) = \frac{pl(B \cup \overline{A}) - pl(\overline{A})}{pl(\Omega) - pl(\overline{A})} \quad \forall B \subseteq \Omega$$

These relations are described under the closed-world assumption. Their extensions under open-world assumption are obtained by suppressing the denominators in $m_A$, $bel_A$ and $pl_A$. The geometrical rule of conditioning has been discussed in Suppes and Zanotti (1977) and Shafer (1976b).

**Conditioning C.5. Specialization.**

Kruse (1990) has studied a conditioning rule that applies to the transferable belief model and is probably the most general form of conditioning that ties in with the idea of basic belief masses that quantify the part of belief that supports a set A of Ω and cannot support a more specific subset of A through lack of information. The idea is that the basic belief mass given to a set A is distributed among the subsets B of A. This concept was also studied by Yager (1986) and Dubois and Prade (1986) when they introduced the ideas of belief inclusions. These authors work under closed-world assumption. Our presentation is made under the open-world assumption.

For each subset $X \subseteq \Omega$, let c(B, X) $\forall X, B \subseteq \Omega$ be non negative coefficients such that:

$$c(B, X) = 0 \quad \text{if } B \not\subseteq X$$
$$\sum_{B \subseteq X} c(B, X) = 1$$

Then the basic belief masses m* obtained by a specialization based on the c(B, X) coefficients are:

$$m^*(B) = \sum_{B \subseteq X} c(B, X) m(X) \quad \forall B \subseteq \Omega$$

The Dempster and the geometric rule of conditioning are specializations. Suppose we known that the answer to the question about Ω is not in $\overline{A} \subseteq \Omega$.

One obtains the unnormalized Dempster's rule of conditioning when c(B, B∪Y) = 1, $Y \subseteq \overline{A}$. Then:

$$m_A(B) = \sum_{Y \subseteq \Omega} c(B, B \cup Y) m(B \cup Y) \quad \forall B \subseteq A$$

One obtains the unnormalized geometric rule of conditioning if c(B, B) = 1 whenever B⊆A and c(∅, B) = 1 whenever B∩A≠∅ otherwise. Normalized rules of conditioning are obtained by further rescaling.

Yager-Kohlas conditioning is not a specialization as the masses given to subsets of $\overline{A}$ are transferred to A.

*Remark*: the unnormalized Dempster's rule of combination is also a specialization. Suppose there exist two basic belief masses assignment $m_1$ and $m_2$ on Ω. Let $m_{12} = m_1 \oplus m_2$ and let c(X, Z) = $m_1$(X|Z). Then (Smets 1991b):

$$m_{12}(X) = \sum_{Y \cap Z = \emptyset, Y,Z \subseteq A} m_1(X \cup Y) m_2(X \cup Z) =$$
$$= \sum_{Z \subseteq \Omega} m_1(X|Z) m_2(Z)$$

### Conditioning C.6. Imaging.

Lewis (1976) considers that probabilities are given to worlds, so each $\omega_i$ represents a world. If one learns that a set $\overline{A}$ of world is impossible (does not contain the answer), then the probabilities given to the worlds in $\overline{A}$ are transferred to the 'closest' worlds in A.

Suppose the 'closest' relation $n(\omega, A): \Omega \times \Omega \to \Omega$ with $n(\omega, A)$ being the world in A that is the closest to world $\omega$. So $n(\omega, A) \in A$ and $n(\omega, A) = \omega$ if $\omega \in A$, i.e., the worlds in A are the closest to themselves.

After conditioning on A by the imaging method, the probability $P(\omega)$ given to a world $\omega$ is transferred to the world $n(\omega, A)$.

Let $F(\omega_i|\omega_j) = 1$   if $\omega_i = n(\omega_j, A)$
           $= 0$   otherwise
Then $P_A(\omega_i) = \sum_{\omega_j \in \Omega} F(\omega_i|\omega_j) P(\omega_j) \quad \forall \omega_i \in \Omega.$

Note that $P_A(\omega_i) = 0$ if $\omega_i \in \overline{A}$. We use the conditional probability notation for F to enhance the fact that F behaves as a conditional probability function.

Of course the idea of 'closest' world needs to be defined and most of the criticism against this conditioning rule focusses on criticism of the idea of closeness between worlds (see Gardenfors 1988).

Gärdenfors (1988) generalizes Lewis's imaging. He considers that the probability given to a world that is learned to be impossible is distributed among the remaining worlds according to some probability distribution that somehow reflect closeness between worlds. $F(\omega_i|\omega_j)$ will represent the portion of the probability given to world $\omega_j \in \overline{A}$ that is transferred to world $\omega_i \in A$. So let:

$F(\omega_i|\omega_j)$ = 0   if $\omega_i \in \overline{A}$
           = 1   if $\omega_j \in A, \omega_i = \omega_j$
           = 0   if $\omega_j \in A, \omega_i \neq \omega_j$
           $\geq 0$   if $\omega_j \in \overline{A}$ and $\omega_i \in A$

and  $\sum_{\omega_i \in \Omega} F(\omega_i|\omega_j) = 1 \quad \forall \omega_j \in \Omega$

Conditioning on A leads to the relations:

$$P_A(\omega_i) = \sum_{\omega_j \in \Omega} F(\omega_i|\omega_j) P(\omega_j) \quad \forall \omega_i \in \Omega$$

Note that $P_A(\omega_i) = 0$ if $\omega_i \in \overline{A}$.

These relations were described for probability function (under closed-world assumption). They can be generalized in two ways: 1) the requirement $F(\omega_i|\omega_j) = 0$ if $\omega_i \in \overline{A}$ is dropped, and 2) the domain from $\Omega$ to $2^\Omega$ is generalized (one assimilates the basic belief masses to probabilities on the power set $2^\Omega$). Let the relation $F:2^\Omega \times 2^\Omega \to [0, 1]$
with:  $\sum_{B \subseteq \Omega} F(B|X) = 1 \quad \forall X \subseteq \Omega$

After conditioning on 'not in $\overline{A}$', one gets:
$m_A(B) = \sum_{X \subseteq \Omega} F(B|X) m(X) \quad \forall B \subseteq \Omega$

This conditioning relation subsumes all those presented so far.

Specialization is obtained if:
  $F(B|X) = 0 \quad \forall B \cap \overline{A} \neq \emptyset$
Yager-Kohlas conditioning is obtained if:
  $F(B|X)$  = 0   $\forall B \cap \overline{A} \neq \emptyset$
         = 1   $B \subseteq A, X = B \cup Y \, \forall Y \subseteq \overline{A}$
         = 1   $B = A, X \subseteq \overline{A}$

### Conditioning C.7. Upper and Lower Bayesian Conditioning.

Suppose I know only that an unknown probability function P over $\Omega$ belongs to a convex subset $\mathcal{P}$ of the set $\mathbb{P}$ of the probability functions defined on $\Omega$. The upper and lower probabilities $P^*$ and $P_*$ define P uniquely, where
$\forall A \subseteq \Omega \quad P^*(A) = \max \{P(A) : P \in \mathcal{P}\}$
           $P_*(A) = \min \{P(A) : P \in \mathcal{P}\}$.

Suppose one asks for the value of $P(B|A) = \frac{P(A \cap B)}{P(A)}$. All that can be said is that $P(B|A)$ is between the upper and lower conditional probabilities $P^*(B|A)$ and $P_*(B|A)$, where $\forall A \subseteq \Omega$ (Smets 1987):
      $P^*(B|A) = \max \{P(B|A) : P \in \mathcal{P}\}$
      $P_*(B|A) = \min \{P(B|A) : P \in \mathcal{P}\}$.

It can be shown that (see Dempster 1967, Fagin and Halpern 1990, Jaffray 1990, Zhang 1989):
$$P^*(B|A) = \frac{P^*(A \cap B)}{P^*(A \cap B) + P_*(A \cap \overline{B})}$$

$$P_*(B|A) = \frac{P_*(A \cap B)}{P_*(A \cap B) + P^*(A \cap \overline{B})}$$

Fagin and Halpern (1990), Jaffray (1990) and Zhang (1989) show that if $P_*(.)$ is a belief function, then $P_*(.|A)$ is also a belief function. Jaffray (1990) also provides the basic belief masses derived from $P_*(.|A)$ by the inverse Moebius transform.

It can be shown that the Upper and Lower Bayesian Conditioning is not a specialization. It is a generalized imaging where the coefficents depend on the basic belief masses of the initial belief function.

## 2. THE SCENARIO: THE VOTING INTENTIONS STUDY.

To illustrate the meaning of the various rules of conditioning we have described, we present a scenario that deals with objective data, induced objective proportions, and where the various forms of conditioning can be described, depending on the contextual information.

Suppose I organize a study on how people will vote in the next election. Let $\Omega$ = {a, b, c, d, e} be the set of candidates. One candidate must be selected by 100 voters. Each voter may vote for only one candidate. Voting will be next Sunday and today, Monday, I shall ask each potential voter to indicate for whom he intends to vote. But voters' opinions, today, are not firmly established and some voters can only point to a subset A of $\Omega$ that contains the name of the candidate they will vote for, but they have not yet decided definitively among these candidates in set A. We accept that voters will always vote for one of the candidates belonging to the set they provided to on Monday; opinions can only be made more specific.

| Sets Answered | Frequencies | Sets after Conditioning |
|---|---|---|
| {a} | 5 | ? |
| {a, b} | 8 | ? |
| {a, b, c} | 15 | {c} |
| {b, c, d} | 21 | {c, d} |
| {a, b, c, d} | 29 | {c, d} |
| {d, e} | 22 | {d, e} |

Table 1. Distribution of voters' intentions Monday, and sets of candidates that remain after the conditioning on {c, d, e}. ? indicates ambiguities that are discussed in the various conditioning schemes.

The voting intentions of the 100 voters are summarized in table 1. Note that the data do not result from a survey based on a sample, but from an exhaustive study of the whole population. This prevents problems related to sampling variations.

On Sunday, the 100 voters will vote and their votes will generate a frequency distribution over $\Omega$. Let $\mathbb{P}$ be the set of frequency distributions Prop over $\Omega$, where Prop(a) for $a \in \Omega$ is the proportion of voters who vote a, and Prop(A) = $\sum_{a \in A}$ Prop(a) for $A \subseteq \Omega^1$. We will neglect the fact that the population is finite, and accept that Prop(A) for $A \subseteq \Omega$ can take any value in [0, 1].

On Sunday, one element of $\mathbb{P}$ will be selected, the one corresponding to the distribution of votes. But today we do not for instance know Prop(a). We know that Prop(a) is at least 5%, at most 78%. Any value in between is acceptable. So all the Monday data says is that Prop belongs to a subset $\mathcal{P}$ of $\mathbb{P}$ where $\mathcal{P}$ contains all those frequency distributions on $\Omega$ compatible with the observed frequencies given in table 1. The set $\mathcal{P}$ is uniquely defined by the upper and lower proportions Prop* and Prop* where $\forall A \subseteq \Omega$

Prop*(A) = max{ Prop(A) : Prop $\in \mathcal{P}$ }
and   Prop*(A) = min { Prop(A) : Prop $\in \mathcal{P}$ }

Some upper and lower proportions induced by the data of table 1 are given in table 2. The proportions Prop(A) will be such that $\forall A \subseteq \Omega$:

Prop*(A) ≤ Prop(A) ≤ Prop*(A)

| Set | Prop* | Prop* |
|---|---|---|
| {a} | 5 = 5% | 5+8+15+29 = 57% |
| {a, b} | 5+8 = 13% | 5+8+15+21+29 = 78% |
| {a, b, c} | 5+8+15 = 28% | 5+8+15+21+29 = 78% |
| {c} | 0 = 0% | 15+21+29 = 65% |
| {d} | 0 = 0% | 21+29+22 = 72% |
| {c, d} | 0 = 0% | 15+21+29+22 = 87% |
| {c, d, e} | 22 = 22% | 15+21+29+22 = 87% |

Table 2. Values for some upper and lower proportions induced by the data of table 1.

We are facing a typical case of upper and lower proportions generated by random sets.

## 3. CONDITIONING.

I collected my data on Monday, but on Tuesday I learn candidates a and b were killed in a car accident during the night. What can I say about Sunday's elections results?

---

[1] We speak of proportions, not of probablity, in order to avoid any confusion. Probability admits many definitions. Using proportions, and later beliefs, will prevent - hopefully - any confusion.

The 15 voters who answered {a, b, c} will have to vote for c as c is the only remaining candidate and these 15 voters had indicated their willingness to consider a candidate in {a, b, c}. Identically, the 21 and 29 voters who had answered {b, c, d} or {a, b, c, d} will vote for a candidate in {c, d}.

The impact of the conditioning information, a and b are dead, results in a transfer of frequencies similar to the one described in the transferable belief model. The frequency given to a set X is transferred to the set X∩Y once we know that the truth is not in $\overline{Y}$. But what about the 13 voters who answered {a} or {a, b}?

Several scenarios can be considered that lead to the different solutions we wish to illustrate. Some may appear somewhat artificial, but that is not the point. We only wish to give a meaning to each conditioning rule.

The labels of the scenarios are those of the conditioning rules described in section 2.

### Scenario C1 : Compulsory Voting.

Suppose we are in Belgium where voting is compulsory (absentees are fined). The 13 voters must cast their votes. If blank voting is allowed, we might consider they will cast blank votes. Table C.1 presents some upper and lower proportions induced after conditioning on {c, d, e}. Note that 13 blank votes will be collected, so Prop*(∅) = Prop*(∅) = 13%.

| Set | Prop* | Prop* |
|---|---|---|
| {c} | 15 = 15% | 15+21+29 = 65% |
| {d} | 0 = 0% | 21+29+22 = 72% |
| {c, d} | 15+21+29 = 65% | 15+21+29+22 = 87% |
| {c, d, e} | 65+22 = 87% | 15+21+29+22 = 87% |

Table C.1. Values for some upper and lower proportions induced by the frequencies of table 1 and the conditioning C.1. (65 = 15+21+29)

This conditioning rule is similar to Dempster's rule of conditioning under open world assumption.

### Scenario C.2. Free Voting.

Same as Scenario C.1, but we compute proportions among those who do not abstain. This is also the French situation where there is no obligation to vote, in which case the 13 voters would not vote on Sunday. Table C.2 presents some upper and lower proportions induced after conditioning on {c, d, e}

This conditioning rule is similar to Dempster's rule of conditioning under closed-world assumption. It is identical with the C.1 conditioning rule except for the normalization factor (the division by 87).

| Set | Prop* | Prop* |
|---|---|---|
| {c} | $\frac{15}{87} = 17.2\%$ | $\frac{15+21+29}{87} = 74.7\%$ |
| {d} | $\frac{0}{87} = 0.0\%$ | $\frac{21+29+22}{87} = 82.8\%$ |
| {c, d} | $\frac{65}{87} = 74.7\%$ | $\frac{65+22}{87} = 100\%$ |
| {c, d, e} | $\frac{65+22}{87} = 100\%$ | $\frac{65+22}{87} = 100\%$ |

Table C.2. Values for some upper and lower proportions induced by the frequencies of table 1 and the conditioning C.2.

### Scenario C.3. Compulsory Choice.

Back to Belgium context C.1 (compulsory voting), but blank votes are not allowed. So all I know about the 13 voters that pose a problem is that they will vote for a candidate in {c, d, e}. Table C.3 presents some upper and lower proportions induced after conditioning on {c, d, e}.

| Set | Prop* | Prop* |
|---|---|---|
| {c} | 15=15% | 15+21+29+13=78% |
| {d} | 0=0% | 21+29+22+13=85% |
| {c, d} | 15+21+29=65% | 65+22+13=100% |
| {c, d, e} | 65+22+13=100% | 65+22+13=100% |

Table C.3. Values for some upper and lower proportions induced by the frequencies of table 1 and the conditioning C.3.

This conditioning is similar to Yager-Kohlas rule.

### Scenario C.4: Geometrical Rule.

Suppose that each voter who had given an answer that contained a or b is so depressed that he commits suicide. Then only 22 voters are left. The proportions among those who had answered a subset of {c, d, e} will be proportionally rescaled. Table C.4 presents some upper and lower proportions induced after conditioning on {c, d, e}.

| Set | Prop* | Prop* |
|---|---|---|
| {c, d} | 0 = 0% | $\frac{22}{22} = 100\%$ |
| {d, e} | $\frac{22}{22} = 100\%$ | $\frac{22}{22} = 100\%$ |

Table C.4. Values for some upper and lower proportions induced by the frequencies of table 1 and the conditioning C.4.

This conditioning corresponds to the geometrical rule of conditioning, a form also encountered naturally in scenarios based on random sets (Smets 1990a)

### Scenario C.5: Specialization

Specialization is obtained when I consider that the information "a and b are dead' allows me to reconsider each voter's answer. For each type of answer, I consider that those who select it will be distributed among its subsets not containing a and b according to a known distribution. The coefficients c of the specialization are these probabilities. As an example, suppose I know (by my knowledge of the political links among the candidates) that: 1) among those who answered {b, c, d}, one third will vote c, one third will vote d, one third is still undecided, 2) among those who answered {a, b, c, d}, half will vote d, the other half is still undecided, and 3) among those who answer {d, e}, half will vote d, the others will vote e. The 13 voters who answer {a} and {a, b} will cast blank votes as in scenario C.1 (normalization can be introduced as in scenario C2). Table C.5 presents some upper and lower proportions induced after conditioning on {c, d, e}.

| Set | Prop* | Prop* |
|---|---|---|
| {c} | 15+7=22% | 15+14+14.5=43.5% |
| {d} | 7+14.5+11=32.5% | 14+29+11=54% |
| {c, d} | 65+11=76% | 65+11=76% |
| {c, d, e} | 65+22=87% | 65+22=87% |

**Table C.5.** Values for some upper and lower proportions induced by the frequencies of table 1 and the conditioning C.5.

This conditioning form corresponds to the specialization.

### Scenario C.6.1. Introspective Realloca-tion: level 1.

Maybe I know more about the candidates than their names. I know that a and b had similar political orientations and that among c, d and e, c is politically the closest to a and b. Then I might consider that the 13 voters without a candidate will vote for c. Table C.6.1 presents some upper and lower proportions induced after conditioning on {c, d, e}

| Set | Prop* | Prop* |
|---|---|---|
| {c} | 15+13=28% | 15+21+29+13=78% |
| {d} | 0=0% | 21+29+22=72% |
| {c, d} | 65+13=78% | 65+22+13=100% |
| {c, d, e} | 65+22+13=100% | 65+22+13=100% |

**Table C.6.1.** Values for some upper and lower proportions induced by the frequencies of table 1 and the conditioning C.6.1.

This conditioning generalizes the imaging conditioning introduced by Lewis where:

$$\forall Y \subseteq \overline{A} \quad F(B|B \cup Y) = 1 \quad B \subseteq A$$
$$= 0 \quad \text{otherwise}$$

### Scenario C.6.2. Introspective Realloca-ion: level 2.

Generalizing Scenario C.6.1, we might consider c as politically very close to a and b, d as quite similar to a and b, and e totally different. So we might accept that there is a certain proportion of the 13 voters that will vote for c and the remainder will vote for c or d. Let the proportion of people that will vote for c given they had decided to vote for a or b be 0.4, and the others will vote for {c, d}.

Table C.6.2 presents some upper and lower proportions induced after conditioning on {c, d, e}.

| Set | Prop* | Prop* |
|---|---|---|
| {c} | 15+5.2=20.2% | 15+21+29+13=78% |
| {d} | 0=0% | 21+29+22+7.8=80% |
| {c, d} | 65+13=78% | 65+22+13=100% |
| {c, d, e} | 65+22+13=100% | 65+22+13=100% |

**Table C.6.2.** Values for some upper and lower proportions induced by the frequencies of table 1 and the conditioning C.6.2.

This form of conditioning generalizes Gärdenfors's Imaging to power sets, where

$$\forall Y \subseteq \overline{A} \quad F(B|Y) \geq 0 \quad B \subseteq A$$
$$= 0 \quad \text{otherwise}$$

### Scenario C.6.3. Introspective Realloca-tion: level 3.

You might be even more subtle in your opinion than in scenario C.6.2. You might consider differently the 5 voters who answered {a} and the 8 who answered {a, b}.

| Set | Prop* | Prop* |
|---|---|---|
| {c} | 15+6=21% | 65+13=78% |
| {d} | 0=0% | 21+29+22+5=77% |
| {c, d} | 65+11=76% | 65+22+13=100% |
| {c, d, e} | 65+22+13=100% | 65+22+13=100% |

**Table C.6.3.** Values for some upper and lower proportions induced by the frequencies of table 1 and the conditioning C.6.3.

They are not identical and you might consider that the distribution of the 5 among {c, d, e} is different from the distribution of the 8. For instance, you might consider that 1) among the 5 voters who answered {a} Monday, 40% (=2) would answer {c} Tuesday, the other (=3) {c, d}, and 2) among the 8 voters who answered {a, b}, half (=4) would answer {c}, one quarter (=2) would answer {c, d} and the last quarter (=2) would answer {c, e}. Table C.6.3 presents some upper and lower proportions induced after conditioning on {c, d, e}.

This corresponds to the generalization of the imaging where

$$F(B|B) = 1 \quad \text{if } B \subseteq A$$
$$F(B|X) \geq 0 \quad \text{if } B \subseteq A, X \subseteq \overline{A}$$
$$= 0 \quad \text{otherwise}$$

### Scenario C.7. Upper and Lower Bayesian Conditioning.

Before learning that a and b are dead, i.e., Monday evening, I would like to assess the proportion of those who will vote for c among those who will vote for c or d or e on Sunday. Should I know the frequency distribution of the Sunday votes, I would compute

$$\text{Prop}(\{c\}|\{c, d, e\}) = \frac{\text{Prop}(\{c\})}{\text{Prop}(\{c, d, e\})}$$

But I do not know these values. All I know are upper and lower limits for each Prop. I know that Prop is in a subset $\mathcal{P}$ of $\mathbb{P}$, the set of frequency distributions on $\Omega$. For each Prop in $\mathcal{P}$, I can compute Prop({c}|{c, d, e}).

| Set | Prop* | Prop* |
|---|---|---|
| {c} | $\frac{0}{0+21+29+22} = 0\%$ | $\frac{65}{65+0} = 100\%$ |
| {d} | $\frac{0}{0+65+22} = 0\%$ | $\frac{21+29+22}{21+29+22+0} = 100\%$ |
| {c, d} | $\frac{0}{0+22} = 0\%$ | $\frac{65+22}{65+22+0} = 100\%$ |
| {d, e} | $\frac{22}{22+65} = 25.3\%$ | $\frac{21+29+22}{21+29+22+0} = 100\%$ |

**Table C.7.** Values for some upper and lower conditional proportions induced by the frequencies of table 1 and the conditioning C.7.

Consequently I know that the upper lower limits of Prop({c}|{c, d, e}) are between the upper and lower conditional proportions Prop*(.|{c, d, e}) and Prop*(.|{c, d, e}) where

$$\text{Prop}^*(A|\{c,d,e\}) = \max\{\frac{\text{Prop}(\{c\})}{\text{Prop}(\{c, d, e\})} : \text{Prop} \in \mathcal{P}\}$$
$$= \frac{\text{Prop}^*(\{c\})}{\text{Prop}^*(\{c\}) + \text{Prop}_*(\{d, e\})}$$

$$\text{Prop}_*(A|\{c,d,e\}) = \min\{\frac{\text{Prop}(\{c\})}{\text{Prop}(\{c, d, e\})} : \text{Prop} \in \mathcal{P}\}$$
$$= \frac{\text{Prop}_*(\{c\})}{\text{Prop}_*(\{c\}) + \text{Prop}^*(\{d, e\})}$$

This conditioning corresponds to the upper and lower bayesian conditioning.

## 4. BELIEFS INDUCED BY THE PROPORTIONS.

Suppose a voter is going to be selected randomly among the 100 voters (with equiprobability for each voter). The question is to bet on who is the Sunday candidate of this randomly selected voter.

Given the available data, all I can say is that the proportion Prop(A) of voters who will vote for a candidate in set $A \subseteq \Omega$ on Sunday is included between Prop*(A) and Prop*(A). So the probability P(A) that the selected voters will vote for a candidate in A is included between Prop*(A) and Prop*(A). Thus I can build upper and lower probabilities P*(A) and P*(A) for P(A) where P*(A) = Prop*(A) and P*(A) = Prop*(A). Given this set of upper and lower probabilities, I can build the pignistic probabilities BetP of the fact that the randomly selected voter will vote for a candidate in A (Smets 1990b), (see also Smets (1991b) for a practical justification of the pignistic transformation).

$$\text{BetP}(A) = \sum_{X \subseteq \Omega} m(X) \frac{|X \cap A|}{|X|}$$

where |X| is the number of candidates in set $X \subseteq \Omega$. BetP is a probability function, and all bets on $\Omega$ are built on it.

But the knowledge of these upper and lower probabilities can also induce a belief in me about the candidate for which the randomly selected voter will vote. As shown in Smets (1991a), the belief function that quantifies my belief about $\Omega$ is numerically equal to the lower probabilities function (because mathematically the lower probability happens to be a belief function in the present scenario). This is based on the maximal-minimal isopignistic transformation described in Smets (1991a). The pignistic transformation of this belief function is of course the same as the one derived from the upper and lower probabilities. So analysing the problem directly from the upper and lower probabilities point of view or through the belief function induced by these upper and lower probabilities leads to the same results.

A bet on who will be the Sunday winner is not analysed. It requires a study of our belief concerning the subsets of $\mathbb{P}$. The belief that x is the winner is equal to the belief allocated to those frequency distributions in $\mathcal{P} \subseteq \mathbb{P}$ where x is the most frequent observation. That is a completely different problem altogether and will not be dealt with.

## 5. CONCLUSIONS.

We have shown that conditioning can be performed by many rules, and through an illustrative example, we have provided scenarios that lead to each form of conditioning. This study is not exhaustive as other forms of updating can - and have been - suggested (Cano and Moral 1990, Moral and De Campos 1990). We hope that these illustrative examples will help the user understand the meaning of the various conditioning rules we have studied.

### Acknowledgements.

The following text presents some research results of the Belgian National incentive-programme for fundamental research in artificial intelligence initiated by the Belgian State, Prime Minister's Office, Science Policy Programming. Scientific responsibility is assumed by the author. Research work has been partly supported by the DRUMS project which is funded by a grant from the Commission of the European Communities under the ESPRIT II-Program, Basic Research Project 3085.

### Bibliography.

CANO J.E. and MORAL S. (1990) Combination of incomplete probabilistic information. (unpublished manuscript)

DEGROOT M.H. (1970) Optimal statistical decisions. McGraw-Hill, New York.

DEMPSTER A.P. (1967) Upper and lower probabilities induced by a multplevalued mapping. Ann. Math. Statistics 38: 325-339.

DUBOIS D. and PRADE H. (1986) A set theoretical view of belief functions. Int. J. Gen. Systems, 12:193-226.

FAGIN R. and HALPERN J. (1990) A new approach to uptdating beliefs. 6th Conf. on Uncertainty in AI.

GARDENFORS P. (1988) Knowledge in flux. Modelling the dynamics of epistemùic states. MIT Press, Cambridge, Mass.

JAFFREY J.Y. (1990) Bayesian conditioning and belief functions. IPMU-1990.

KOHLAS J. (1989) The reliability of reasoning with unreliable arguments. Inst. Automation and Oper. Research, Univ. Freiburg, Technical report 168.

KRUSE R. and SCHWECKE E. (1990) Specialization: a new concept for uncertainty handling with belief functions. Int. J. Gen. Systems (to appear)

LEWIS D. (1976) Probabilities of conditionals and conditional probabilities. Philosophical Review 85: 297-315.

MORAL S. and DE CAMPOS L.M. (1990) Updating uncertain information. (unpublished manuscript)

PLANCHET B. (1989) Credibility and conditioning. J. Theor. Probabil. 2:289-299.

SHAFER G. (1976a) A mathematical theory of evidence. Princeton Univ. Press. Princeton, NJ.

SHAFER G. (1976b) A theory of statistical evidence. in Foundations of probability theory, statistical inference, and statistical theories of science. Harper and Hooker ed. Reidel, Doordrecht-Holland.

SMETS Ph. (1988) Belief functions. in SMETS Ph, MAMDANI A., DUBOIS D. and PRADE H. ed. Non standard logics for automated reasoning. Academic Press, London p 253-286.

SMETS P. (1987) Upper and lower probability functions versus belief functions. Proc. International Symposium on Fuzzy Systems and Knowledge Engineering, Guangzhou, China, July 10-16, pg 17-21.

SMETS Ph. (1990a) The transferable belief model and random sets. To appear in Int. J. Intell. Systems.

SMETS Ph. (1990b) Construucting the pignistic probability function in a context of uncertainty. Uncertainty in Artificial Intelligence 5, Henrion M., Shachter R.D., Kanal L.N. and Lemmer J.F. eds, North Holland, Amsterdam, , 29-40.

SMETS Ph. (1991a)Belief induced by the knowledge of some probabilities. Submitted for publication.

SMETS Ph. (1991b) Belief functions: the disjunctive rule of combination and the generalized Bayesian theorem. (submitted for publication)

SUPPES P. and ZANOTTI M. (1977) On using random relations to generate upper and lower probabilities. Synthesis 36:427-440.

YAGER R.R. (1985) On the Dempster-Shafer framework and new combination rules. Technical report MII-504, Machine Intelligence Institute, Iona College.

YAGER R.R. (1986) The entailment principle for Dempster-Shafer granules. Int. J. Intell. Systems 1:247-262

ZHANG L (1989) A new proof to theorem 3.2 of Fagin and Halpern's paper. Unpublisshed Memorandum, University of Kansas, Business School.